\documentclass{article}

\usepackage{arxiv}

\usepackage[utf8]{inputenc} 
\usepackage[T1]{fontenc}    
\usepackage{url}            
\usepackage{booktabs}       
\usepackage{amsfonts}       
\usepackage{nicefrac}       
\usepackage{microtype}      
\DisableLigatures[f]{encoding = *, family = * }
\usepackage{amsmath,amssymb}

\usepackage{hyperref}
\usepackage{nameref}

\usepackage{graphicx}
\usepackage[table]{xcolor}
\usepackage{epstopdf}

\usepackage{array}

\newcolumntype{+}{!{\vrule width 2pt}}

\newlength\savedwidth

\newcommand\thickhline{\noalign{\global\savedwidth\arrayrulewidth\global\arrayrulewidth 2pt}%
\hline
\noalign{\global\arrayrulewidth\savedwidth}}

\DeclareMathOperator*{\argmax}{argmax}
\newcommand{\pvec}[1]{\vec{#1}\mkern2mu\vphantom{#1}}

\title{Learning efficient haptic shape exploration with a rigid tactile sensor array}

\author{
  Sascha Fleer\thanks{These authors contributed equally to this work.} \\
  Neuroinformatics Group \\
  Bielefeld University \\
  Germany \\
  \texttt{sfleer@techfak.uni-bielefeld.de} \\
   \And
   Alexandra $\textrm{Moringen}^*$\\
  Neuroinformatics Group \\
  Bielefeld University \\
  Germany \\
  \texttt{abarch@techfak.uni-bielefeld.de} \\
   \And
   Roberta L. Klatzky \\
   Department of Psychology \\
   Carnegie Mellon  University \\
   USA \\
   \And
    Helge Ritter \\
    Neuroinformatics Group \\
    Bielefeld University \\
    Germany \\
}

\begin{document}
\maketitle

\section*{Abstract}
Haptic exploration is a key skill for both robots and humans to
discriminate and handle unknown objects or to recognize familiar objects.  Its
active nature is evident in humans who from early on
reliably acquire sophisticated sensory-motor capabilities for active
exploratory touch and directed manual exploration that associates
surfaces and object properties with their spatial locations. This is in  stark
contrast to robotics. In this field, the relative lack of good
real-world interaction models --- along with very restricted sensors and a scarcity
of suitable training data to leverage machine learning methods --- has so far
rendered haptic exploration a largely underdeveloped skill.
In robot vision however, deep learning approaches and an abundance of
available training data have triggered huge advances.

In the present work, we connect recent advances in \emph{recurrent
  models of visual attention} with previous insights about the
organisation of human haptic search behavior, \emph{exploratory
  procedures} and \emph{haptic glances} for a novel architecture that
learns a generative model of haptic exploration in a simulated
three-dimensional environment. This environment contains a set of
rigid static objects representing a selection of one-dimensional local
shape features embedded in a 3D
space: an edge, a flat and a convex surface.
The proposed algorithm simultaneously optimizes main perception-action
loop components: feature extraction, integration of features over
time, and the control strategy, while continuously acquiring data
online.  Inspired by the \emph{Recurrent Attention Model}, we
formalize the target task of haptic object identification in a
reinforcement learning framework and reward the learner in the case of
success only. We perform a multi-module neural network training,
including a feature extractor and a recurrent neural network module aiding pose
control for storing and combining sequential sensory data.
The resulting haptic meta-controller for the rigid
$16\times16$ tactile sensor array moving in a physics-driven
simulation environment, called the \emph{Haptic Attention Model}, performs a sequence of haptic glances, and
outputs corresponding force measurements.  The resulting method has
been successfully tested with four different objects. It achieved
results close to $100 \%$ while performing object contour exploration
that has been optimized for its own sensor morphology.

\section*{Introduction}

While the sense of touch is central to human life, tactile capabilities of robots are currently
hardly developed. This stark contrast becomes even more apparent if one compares touch and vision:
while good camera sensors have become affordable and ubiquitous items and huge image and video
databases together with deep learning have brought computer vision close (some would argue on par)
to human vision \cite{SzegedyLJSRAEVR14, HeZRS15, Levine:2017cu},
comparable advances in robot touch are widely lacking \cite{Okamura2001, MartinsFD14, Tian2019MBF, Lee2019MSV}.

One reason is the very limited maturity of tactile sensors as 
compared with human skin.  A second and deeper reason is that touch
differs from vision in an important way: while looking at an object
leaves its state unaffected, touch requires physical contact, coupling
the sensor and the object in potentially complex and rich ways that
usually also change the position, orientation or even the shape of the
object.  Human haptics makes active and sophisticated use of this
richness to lend us skills such as haptic exploration, discrimination,
manipulation and more.  Large parts of these tasks are hard or
impossible to model sufficiently accurately to replicate them on
robots, thereby calling again for machine learning approaches similar
to those that were highly successful in vision. However, the highly
interactive nature of touch makes not only the learning problem itself
much more difficult but also creates a problem for the availability of
meaningful training data, since information about interactive haptics
is much harder to capture in databases of static tactile patterns.  As
a consequence, learning approaches for the modality of interactive
touch are still largely in their infancy and tactile skills
enabling robots to establish and control rich and safe contact with
objects or even humans are still a largely unsolved challenge which
severely limits the use of robots in both domestic and industrial
applications.

In this work we focus on using machine learning for the synthesis of one central and important haptic skill:
the discrimination of unknown object shapes through a sequence of actively controlled haptic contacts between a sensor and the object surface.
Our approach builds on recent advances that show how a deep network can be made to learn to integrate a sequence of visual observations to discriminate visual patterns.
We extend this approach from the visual to the haptic domain and --- by taking inspiration
from insights about the organization of haptical exploration in humans --- we create a potentially interesting new bridge between a computational understanding of interactive touch in robotics and in human haptics.

In humans, haptic capabilities are available at birth, for example,
those that are necessary for a neonate to nurse. Over the course of early development, increasingly
sophisticated haptic exploration comes on-line, as children acquire motor control and the ability to focus attention.
By pre-school age, children demonstrate adult-like patterns of exploration \cite{Kalagher2011YCH}
that they gate according to contextual demands \cite{Klatzky:2005ty}.
This developmental process results in a small set of optimized action patterns, widely known under the term
\textit{exploratory procedures} (EPs) \cite{Klatzky1987TMT}.
Humans use EPs to extract properties such as texture, hardness,
weight,  volume, or local shape features.

Under some
circumstances, the level of complexity in haptic exploration can be
effectively reduced to what was termed the
\textit{haptic glance} by Klatzky and Ledermann \cite{Klatzky1995}.
Specifically, they define the haptic glance as brief,
spatially constrained contact that involves little or no movement of
the fingers.  In the same work they pose the question how the
information from a haptic glance is translated into effective
manipulation.  Following this work, we are interested in a
connection/transition between a haptic glance and an exploratory
procedure.  We propose that a haptic glance constitutes an atomic,
primitive exploratory entity. We furthermore assume that an EP can be
represented by a sequence of such primitives, if parameterization of
each individual haptic glance is chosen in an optimal way.  On a
long-term scale, we are targeting the question:
How can one model optimal control of haptic glances for optimal task-specific
haptic exploration of an unknown object or scene?
Will the resulting
sequence of haptic glances emerge as a full EP? In order to answer
this question affirmatively, such a control model should ideally
contain a strategy to efficiently extract task-specific cues based on
previously available information (if any), and integrate these over
time.  For computational purposes we make the following assumptions.
Firstly, we assume that a haptic glance --- being the simplest
haptically directed action --- is a foundation for any more complex
haptic behavior, including haptic exploratory procedures of any
type. Therefore, it is our goal to learn an optimal sequence of haptic
glances, adapted to a given task and a sensor morphology that is
provided beforehand and is specific for a given robot
platform. Secondly, we assume that a haptic glance is defined by a
tuple consisting of a pressure profile yielded by the tactile sensor
at contact and the corresponding sensor pose.

Robots, like humans, benefit from haptic sensors in order to find,
identify, and manipulate objects. 
Tactile sensing applications in robotics are built out of two different
categories~\cite{Fishel:2012gp}.
The first one is called ``perception
for action'', which utilizes the tactile information to solve dexterous
manipulation tasks including grasping, slip prevention.  The second
category, which has recently become a popular area of research, is
named ``action for perception'', dealing with recognition and
exploration \cite{Chu:2013fo,CHU2015279,Pape:2012jf}.
Recent developments have added machine learning
techniques in order to learn exploration strategies,
feature extraction or a better estimation of different quantities.
One class of methods  is
\emph{reinforcement learning}, a biologically inspired class of learning methods
in which the agent learns by gathering data through the active exploring of the
environment \cite{Sutton:1998wc}.
It is applied to teach a robot dexterous manipulation \cite{vanHoof2015,Rajeswaran2017}
or to use learned exploration
strategies in the form of tactile skills in order to facilitate
exploration as studied for surface classification
\cite{Pape:2012jf}.

The approach employed in this work provides one possible solution to a
typically puzzling question: how to couple optimization of both
above-mentioned directions, ``perception for action'', and ``action for
perception''. In computer vision, the analogous question has already been investigated by measures of \textit{recurrent models of visual attention}
(RAM) \cite{Mnih:2014ti,Ba:2014ws}.  RAM acquires image glimpses by
controlling the movement of a simulated eye within the image.  The
modeling approach is inspired by the fact that humans are not
perceiving their environment as a whole image. Instead, they see only
parts of the scene, while the location of the fixations depends on
the current task \cite{Hayhoe:dk,Mathe:2013tg}. The model
gathers information about the environment directed by image-based and
task-dependent saliency cues~\cite{Itti:tq,Itti:2001cl}.
Information extracted from these foveal ``glimpses'' is then combined in
order to get an accumulated understanding of the visible scene.
RAM applied to control of the sequences of haptic
glances optimizes both above-mentioned directions simultaneously in a
series of iterative steps, and enables us to find an optimal solution
for a given tactile end-effector, with respect to its own constraints and the
spatio-temporal resolution of the acquired data.

Inspired by this work, we present a
framework that is able to identify four different objects
using a tactile sensor array within a simulated environment. The object
classification and pose control are formalized as a sequential
decision-making process within a reinforcement learning framework,
where an artificial agent is able to perform multiple haptic glances
before the final estimation of the object's class. During the training
of a multi-component deep neural network, we learn how to control the pose
of the rigid tactile sensor in a way that is beneficial for the
classification task. To enable integration of information gained through multiple haptic
glances, we employ a recurrent neural network as one building block of
this architecture.
The next section describes the simulation setup and the employed algorithm, together with the training procedure.
After presenting the conducted experiments, we summarize and discuss the
obtained results.

\section*{Scenario and Experimental Setup}

\begin{figure}[!h]
    \centering{
    \includegraphics[width=\linewidth]{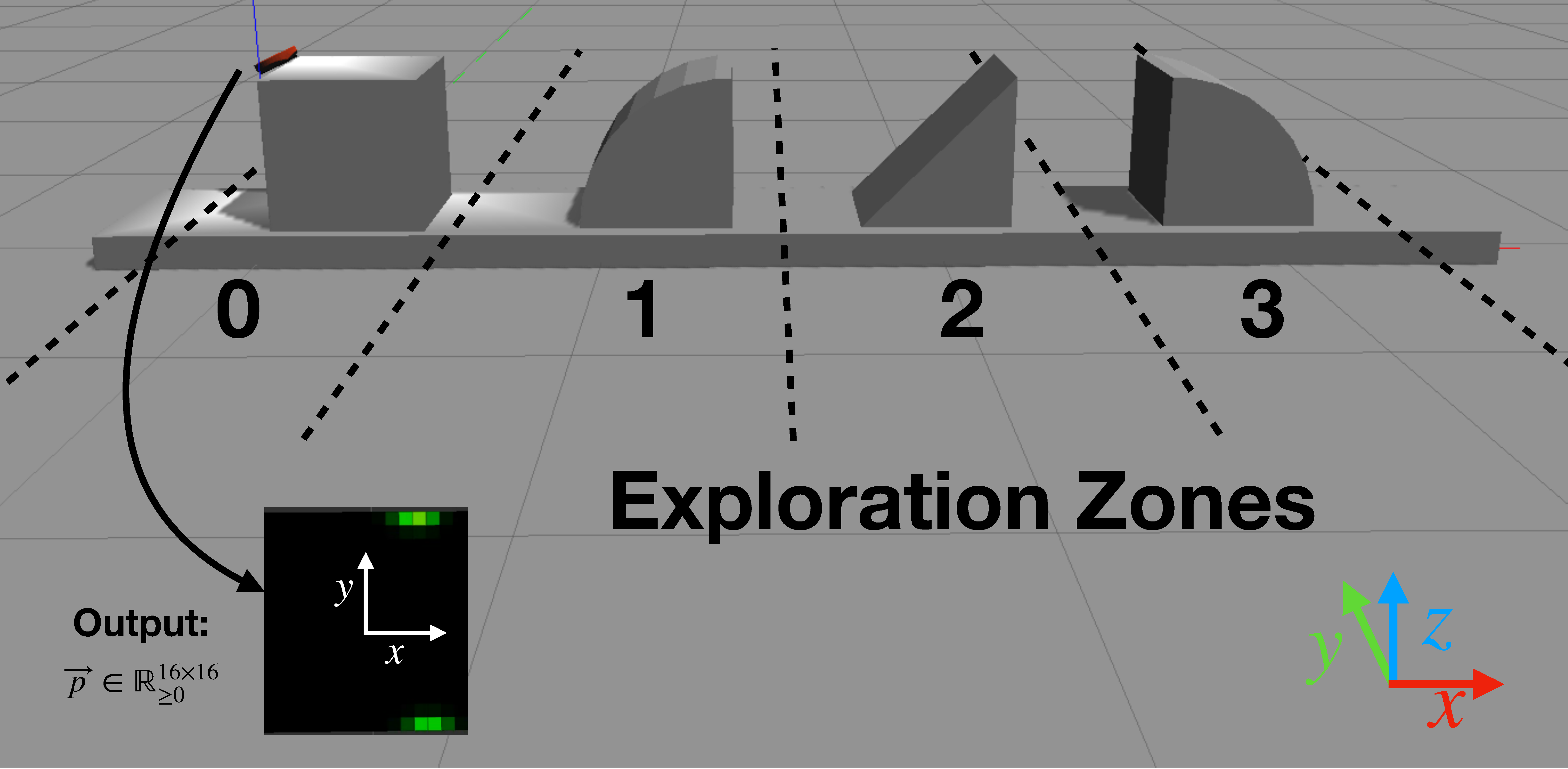}
}
\caption{{\bf Gazebo Simulation} The image displays a view of the
  linear object arrangement along the $x$-axis with four objects from 0 to 3 and the simulated Myrmex sensor (small red
  blob touching object 0). The bottom
  side of the sensor contains the square-shaped pressure sensitive tactile array,
  measuring a pressure profile on contact with the object that is visualized by
  the square on the bottom left of the image.
  Additionally, the borders of the \emph{exploration zone} for each object are indicated by dotted lines.}\label{fig:Experiment_Gazebo}
\end{figure}

To develop an efficient haptic controller that can enable a
robot to identify objects with a sequence of haptic glances, we
perform a comprehensive experimental investigation in \texttt{Gazebo}
(see \nameref{S1_Link_Gazebo}), a physics-driven simulation
environment. The simulation consists of two main parts as illustrated in
Fig~\ref{fig:Experiment_Gazebo}.

\subsection*{The Tactile Sensor Array - Myrmex}
The first part is a floating standalone tactile sensor array, modeled
to resemble the \emph{Myrmex}~\cite{Schurmann:2011fu} sensor in order
to ease the transfer to a real robot in future experiments.  It is
constructed out of a circular end-effector mount (red) with a square
sensitive zone (black).  In simulation, one side of the sensor
contains a square-shaped array of $16 \times 16$ cells covering a surface of
$64 \; \mathrm{cm}^2$, whose values
are computed to approximately resemble the values of the real sensor
array (see \nameref{S2_Link_Myrmex}).  Contacts at collision are
estimated by Gazebo's physics engine \texttt{ODE} according to
inter-penetration of objects (intrinsic compliance) and to default
local surface parameters. An example of the contact information
available in Gazebo and its characteristics are shown in
\nameref{S1_Video}.

\begin{figure}[!h]
    \centering{
    \includegraphics[width=0.65\linewidth]{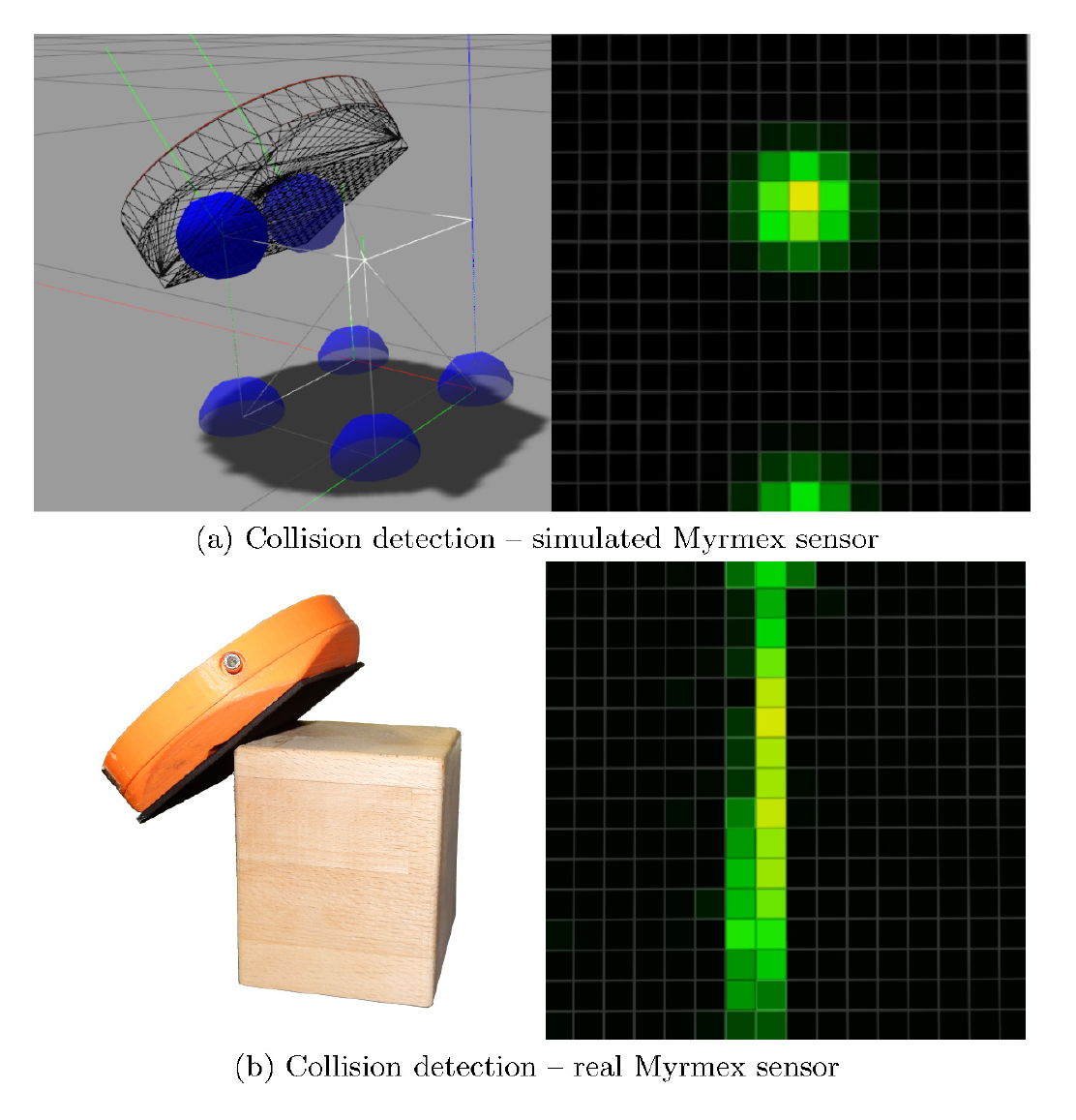}
}
\caption{{\bf Comparison of the simulated and real sensor} A
  comparison of the measurements between simulated and real Myrmex
  sensor when contacting an object edge. The left side of (a) sketches the measuring
  process of the simulated Myrmex. Blue spheres illustrate the measured contact
  information leading to the tactile
  image on the right side of (a). Figure (b)
  shows the real sensor, together with the measured tactile image.
  }\label{fig:comparison_sim_real_myrmex}
\end{figure}

Each contact defined by its position and force vector, generates a
Gaussian distribution around the contact center with amplitude
depending only on the normal force. The standard deviation is
arbitrarily fixed to mimic the deformation of the sensitive foam on
the real sensor.  Mixing the distributions creates a $16 \times 16$
tactile pressure image, that is represented as an array of floating
point values contrary to the real sensor with only 4096 levels of
pressure.  When measuring the collision with an edge as it is
illustrated in Fig~\ref{fig:Experiment_Gazebo}, we expect to see a
line. However, due to the limitations of the collision library,
we acquire the image presented in the bottom left
corner. In Fig~\ref{fig:comparison_sim_real_myrmex} the
tactile image for a contact with a cuboid is shown for both the
simulated Myrmex sensor (Fig~\ref{fig:comparison_sim_real_myrmex}
(a)), and the real sensor (Fig~\ref{fig:comparison_sim_real_myrmex} (b)).
The collision library \texttt{libccd} used by the ODE simulation
engine of Gazebo can generate only two contact points at a time (see
\nameref{S1_Video}).
Consequently, it is not possible to produce an edge in the resulting
tactile image. On the contrary, the real sensor produces a tactile
image in which the expected line of contact is visible.

Communication with the simulated sensor in Gazebo is
performed via a
\texttt{ROS-interface} (see \nameref{S3_Link_Ros}).

\subsection*{Stimulus Material}
The second part is the stimulus material. It exists as a static set of 3D
objects that are
distributed in the simulation environment, but also in the form of
real 3D wooden building blocks with 3D elementary shapes carved on
top. Our current set of elementary shape types consists of
approximately 60 prototypes. A combination of such building blocks
forms the so-called Modular Haptic Stimulus Board (MHSB)
(see \nameref{S2_Video} for design and applications
and \nameref{S1_Project} for the MHSB project web-site).  By
rearranging the blocks, MHSBs of different sizes and different shape
configurations have been previously employed in a range of studies of
haptic exploration and search in humans
(e.g.,~\cite{Moringen2017HSC,Krieger2016SFS,Moringen2016SPD}).
Through its modularity, MHSB enables a flexible experimental design
resulting in a wide range of 3D shape landscapes.

All shapes within the current setup are rigid, stationary and have the
same height.  Building blocks employed for this experiment, $9\times9 \; \mathrm{cm}$ each, were chosen, firstly, to suit the size of the real Myrmex
sensor and the restrictions of its control with the real KUKA robot
arm. For this work, we have chosen a set of objects locally
representing basic types of one-dimensional curvature features,
e.g. edge, flat descendent/horizontal surface, and a convex 
surface.
Due to the fact that concave surfaces may be more challenging for the simulated sensor,
we are omitting them in the current work.
This one-dimensional curvature design enabled us to constrain
parameterization of haptic glances to two dimensions, translation and
rotation along one axis, together with the linear arrangement of the shapes, 
without loss of generality.
In case new
features are considered within the experimental stimulus design, new types of
control parameters as well as new output have to be used for an
implementation of the haptic glance controller.  For example, in
case objects are equal w.r.t. the curvature and can be differentiated
based on height only (a set of cuboids of different heights), a haptic
glance controller needs to output the height of  collision with the
object as well as the pressure profile.

\subsection*{Haptic Control of The Simulated Myrmex}

Haptic control consists of two parts, a low-level controller that
performs haptic glances and a higher-level controller that provides
parameterization for the low-level controller and is responsible for
solving the task. 

\paragraph{The High-Level Meta-Controller --- HAM}
The process of haptic exploration is operated by the so-called meta-controller:
the \emph{Haptic Attention Model} (HAM).
It is represented by a deep neural network and is described in detail in the
Methods Section. Its main task is to classify the given object,
while constantly providing a new expedient target pose
$\xi=(x_g,y_g,z_g,e_1,e_2,e_3)$ of the sensor, including three position
coordinates $(x_g,y_g,z_g)$ and
three orientations $(e_1,e_2,e_3)$, to the low-level controller for further exploration.
It performs the
optimization for the parameterization of haptic glances based on the
state of the networks' working memory, a representation of the
previously acquired haptic data.
For proof of concept, we restricted the number of parameters that have to be
provided by the HAM to the position along the $x$-axis and
the angle around the
$y$-axis.  Before the execution of the haptic glance, the sensor is
positioned at a specific pose where $x_g$ and the Euler angle $e_2$
are specified by the output of the network
$\vec{l}=(x_g,e_2)^\top$. For the sake of readability, the alterable
position $x_g$ is called $x$ and the angle $e_2$ is called $\varphi$
in the following sections.

\begin{figure}[!h]
    \centering{
    \includegraphics[width=\linewidth]{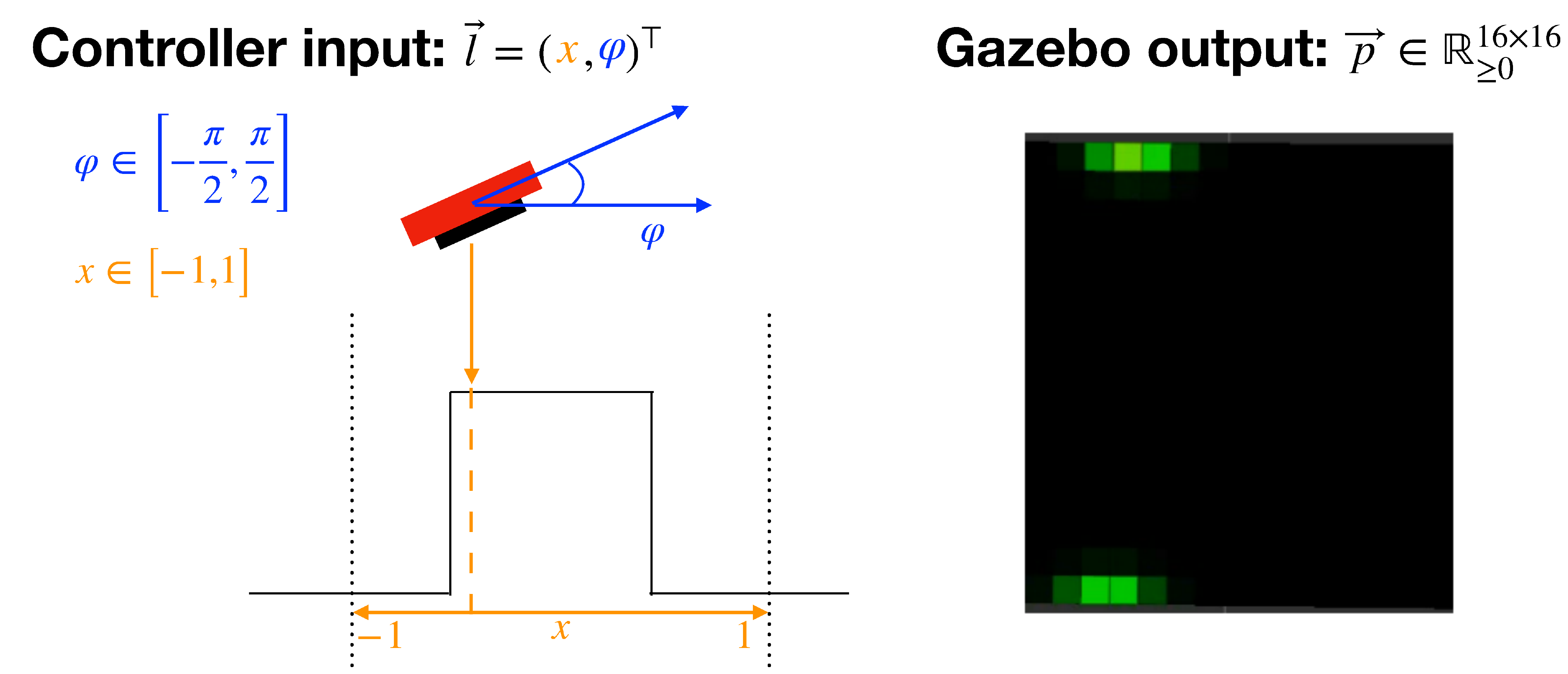}
}
\caption{{\bf Schematic Illustration of the Experiment's Core Idea and Its
    Realization in Simulation} The experimental setup contains four
  objects whose positions are static. Myrmex gathers information
  about the objects by performing haptic glances at position $x$ and
  orientation $\varphi$ around the $y$-axis, leading to a $16 \times
  16$ pressure image.}
\label{fig:Experiment}
\end{figure}

\paragraph{The Low-Level Haptic Glance Controller}
Without loss of generality, we use a simplified and naive representation of the
low-level controller as illustrated in Fig~\ref{fig:Experiment}.
It executes a primitive haptic interaction specified
by two parameters which are provided by the HAM.
Given a pose, it outputs the acquired pressure, $g:
(x, \varphi) \rightarrow \vec{p}$.
 An execution of the
glance controller moves Myrmex from
a predefined $(x,y,z)$-position down along the $z$-axis. To this end,
it gradually decreases its height --- indicated by the value $z$ --- while keeping both the
orientation and the $(x,y)$ position constant until a collision with
an object takes place (see \nameref{S3_Video}).
Upon collision with the object, handled by the physics engine with
minimal penetration when the overall pressure level on the sensor
reaches a certain threshold, the motion stops and the sensor outputs its
readings. To compute it, the forces applied to the $16 \times 16$ sensor cells of the
Myrmex sensor are summed up. The threshold value is reached, when an overall force of 
$2 \; \mathrm{N}$ is distributed over the contact surface of the Myrmex, 
i.e. $2 \; \mathrm{N} / 64 \; \mathrm{cm}^2 = 312.5 \; \mathrm{Pa}$.   
The main feature of this controller, implemented with the
``hand of god'' plugin (see \nameref{S4_Link_HoG}), is the
constant sustainment of the sensor's orientation and the $(x,y)$ position
up to the time of collision. This is realized by switching off the
gravity and continuously holding the sensor pose at a predefined
value against the impact of any impulses. By this means, the
full control of both the pose parameters and the resulting tactile
measurement is guaranteed. Additionally, this restricted implementation resembles
 the movement of the sensor when attached to a robot arm.

In this work, determined by the type of tactile sensing available as
well as the restricted design of the haptic object properties, the
haptic glance controller employed by the network is parameterized only 
by the pose. However, the parameterization may be extended, or,
alternatively, a set of differently parameterized haptic glance
controllers, similar to a functional basis, may be employed by the
network. An example of an extension would be a function $\hat{g}:
(x, \varphi) \rightarrow (\vec{p}, h)$ that maps from the pose to the
tuple containing both the pressure and the corresponding height.  Such
parameterization is necessary in case the stimuli differ in height.
If we further extend the shape complexity from the one-dimensional to
a two-dimensional curvature feature, two orientation parameters
instead of one will account for the data acquisition,
i.e. $\hat{g}: (x, \varphi_x, \varphi_y) \rightarrow (\vec{p}, h)$.

\subsection*{Classification Task}
During training and classification, the agent is always presented with
one out of four objects. It explores the restricted object space with
the sensor by performing a predefined number of haptic glances.
In order to learn an exploration policy that is independent of the
object's pose within the global coordinate system, we introduce
\emph{exploration zones}  illustrated with dashed lines in
Fig~\ref{fig:Experiment_Gazebo}.  Exploration zones are pre-defined regions
with their own local coordinate systems, in which the objects are
placed for exploration.  After specification of the exploration zone,
two out of six pose parameters of the tactile sensor can be modified
by the high-level meta-controller as explained in the previous section.
To preclude learning the absolute position of the object, its coordinates within the
simulation space are mapped to an exploration zone
$x \in \left[-1,1 \right]$, corresponding to the range in which the
output of the neural network lies. Due to the location of the
pressure-sensitive surface on only one side of the Myrmex, rotations
are performed within the range $\varphi \in \left[ -\pi/2,
+\pi/2 \right]$.
Further
rotation will not yield contact information between the object and the
sensor surface.  The acquired pressure information is employed not
only to classify the given object but also to determine the next
position and orientation of the sensor in the next exploration step.

\section*{Methods}
\label{sec:methods}

Reinforcement learning is a well-known class of machine learning
algorithms for solving sequential decision-making problems through
maximization of a cumulative scalar-valued reward function
\cite{Sutton:1998wc}.  To formalize our task as a reinforcement
learning problem, the artificial agent receives a reward of $r=1$ for
a correctly classified object and a reward of $r=0$ otherwise.  We
then use the standard formulation of a Markov decision process defined
by the tuple $(S, A, P^A, R, \gamma , S_0)$, where $S$ denotes the set
of states and $A$ the set of admissible actions.  $P^A$ is the set of
transition matrices, one for each action $a\in A$ with matrix elements
$P^a_{\vec{s},\pvec{s}'}$ specifying the probability to end up in state
$\vec{s}'$ after taking action $a$ from state $\vec{s}$.
Finally, $r \in \mathcal{R}
\subset \mathbb{R}$
is a scalar valued reward the agent receives after ending up in $\pvec{s}'$,
$\gamma$ the discount factor and $S_0 \subseteq S$ is the set of
starting states. The goal is to find an optimal policy $\pi:S \rightarrow
A$  that  maximizes the discounted future reward
\begin{equation}
    R_t = \sum \limits_{k=0}^{\infty} \gamma^{k} r_{t+k}.
    \label{eq:total_reward}
\end{equation}
The discount factor $\gamma\in[0,1)$ balances the weighting between present rewards and rewards that
lie increasingly in the future.

A neural network with a set of weights $\theta$ can be employed to solve a reinforcement learning
task, i.e., its output should maximize a given reward function
$R_t$. In this case we can perform a gradient-based policy optimization
with the help of the REINFORCE update rule~\cite{Williams1988,Williams:1992hd}. The general
rule for updating the corresponding weights $\theta$ of
the network is thus given by
\begin{eqnarray}
    \Delta_{\theta} = \alpha \cdot \left[ r_t - b(\vec{s}_t;\theta) \right] \cdot \zeta(\vec{s}_t;\theta),
    \label{eq:reinforce}
\end{eqnarray}
where $\alpha$ defines the learning rate factor, $b$ the reinforcement baseline. $\zeta$ is called the \emph{characteristic eligibility}. It is defined as
\[
  \zeta(\vec{s}_t;\theta) = \frac{\partial \log{f(\vec{s}_t;\theta)}}{\partial \theta},
\]
where  $f(\vec{s}_t;\theta)$ determines the trainable output of the network
as a function of its input $\vec{s}_t$ and its weight parameters $\theta$.
Using REINFORCE, it is thus possible to develop learning rules for stochastic policies that depend
on multiple
input parameters, like an adaptable Gaussian with variable mean $\mu$
and standard deviation $\sigma$. To this end, a neural network is
trained to map the input to a parameterization of the Gaussian
distribution, i.e., $\mu$ and $\sigma$.  Instead of their corresponding
weights $\theta_\mu$ and $\theta_\sigma$, $\mu$ and $\sigma$
themselves can be treated as the adaptable parameters of the Gaussian
$\mathcal{N} (x ;\mu,\sigma)$.  Using this simplification, the
characteristic eligibility for $\mu$ is given by
\begin{eqnarray}
    \zeta_\mu = \frac{\partial \log \mathcal{N} (x ;\mu,\sigma)}{\partial \mu} = \frac{x - \mu}{\sigma^2},
    \label{eq:reinforce_mean}
\end{eqnarray}
where $x$ is the corresponding value, sampled from the Gaussian distribution $\mathcal{N}$.
Analogously, the characteristic eligibility for $\sigma$ is
\begin{eqnarray}
    \zeta_\sigma = \frac{\partial \log \mathcal{N} (x ;\mu,\sigma)}{\partial \sigma} = \frac{(x - \mu)^2 - \sigma^2}{\sigma^3}.
    \label{eq:reinforce_std}
\end{eqnarray}
The details of the application of these equations to our work is
described in the section below.

\subsection*{The Haptic Attention Model}
\label{sec:nn}

\begin{figure}[!h]
    \centering{
    \includegraphics[width=0.75\linewidth]{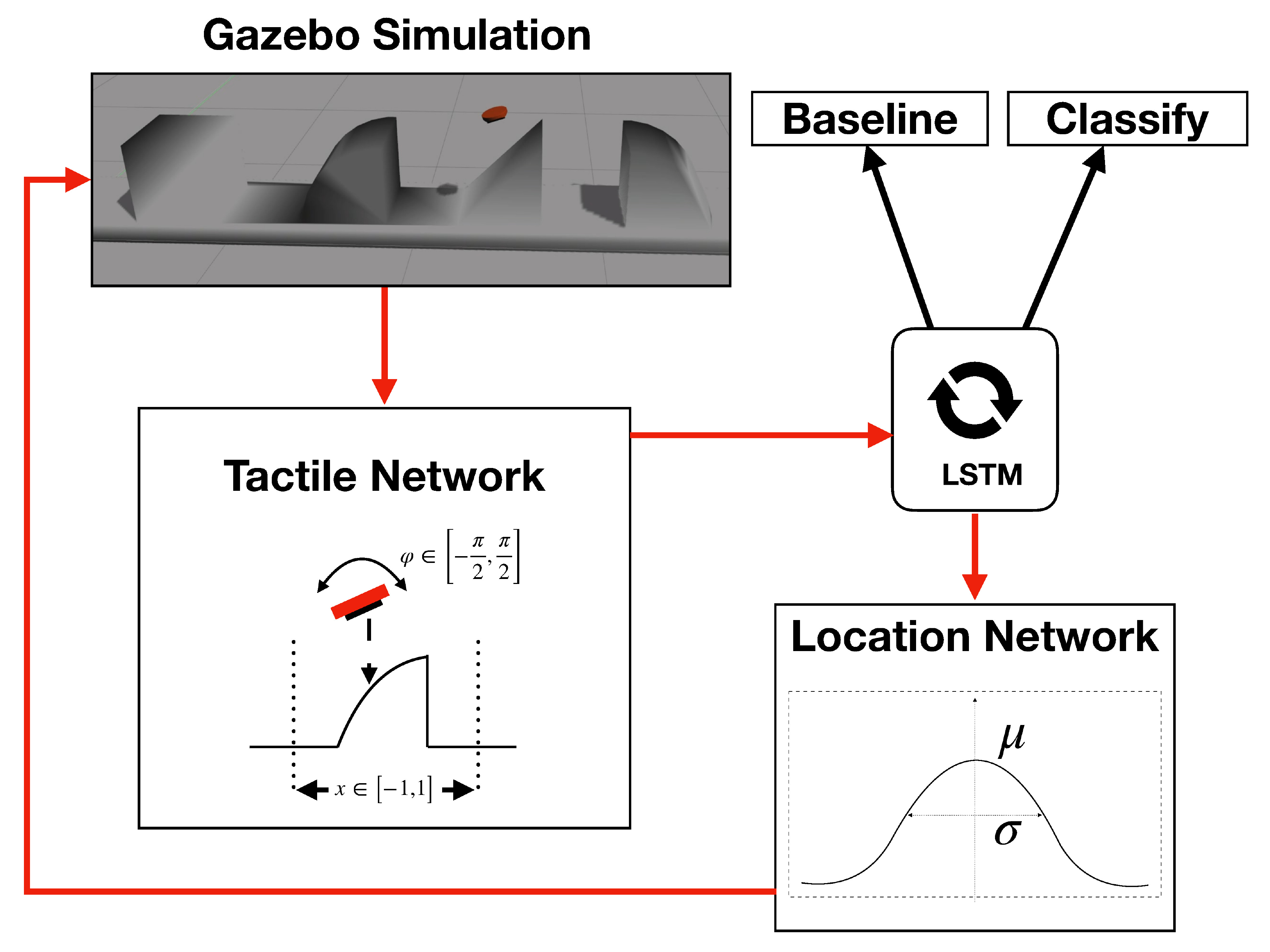}
}
\caption{{\bf Illustration of the used Model} The overall design of the multi-module meta-controller model
 and its interaction with the Gazebo simulation environment.}\label{fig:model}
\end{figure}

In the following, the architecture of our designed high-level meta-controller,
called the \emph{haptic attention model} is described in detail.
An overview of the interaction loop between the network and the simulation
is displayed in Fig~\ref{fig:model}. Inspired by the architectures in
\cite{Mnih:2014ti,Ba:2014ws}, the  meta-controller network  is
constructed from three modules which are described in detail in the
following subsections (See also \nameref{S5_HAM}).
A vector $\vec{s}=(x,\varphi,\vec{p})^\top$ consisting of the sensor
pose $(x,\varphi)$ and the corresponding pressure profile acquired by
Myrmex performing a haptic glance in Gazebo is used as the sensory
input for the network. The $16 \times 16$ pressure matrix is flattened
to a normalized pressure vector $\vec{p}$ with
$\mathrm{dim}(\vec{p})=256$. For the normalization we employ the
$L2$-norm.  Apart from the considerations of numerical stability
during network training (no small/large numbers and no large
differences), the normalization is performed in order to get rid of
artifacts in the data caused by the method chosen to perform a
haptic glance in simulation. These artifacts are specific to moving
the floating Myrmex towards an object at an unknown position in tiny
discrete steps, which is likely to produce a different strength of
signal depending on the distance between the sensor and the object in
the last step prior to collision.  Therefore, normalization is
performed in order to achieve a comparable pressure profile for a
given pose, independent of the force, whose absolute strength in this
particular case is a simulation artifact.

First,
the input is processed through the
\emph{tactile network}, which combines the recorded pressure profile
$\vec{p}$ with its corresponding location $x$ and orientation
$\varphi$ into one single feature vector. The features $\vec{s}$ are then
propagated through a \emph{long short-term memory} (LSTM) network~\cite{Hochreiter:1997fq}.
This kind of neural network belongs to the class of ``recurrent neural
networks'' which have the ability to store, combine and process sequential data.
It is constructed using hidden states of 256 neurons.
The LSTM provides features to the \emph{object classifier} and to the \emph{location
network} that in turn provides a new pose. Although the classification of the object can be done within each glance, we usually refer to the classification result after the
final glance.

If not stated otherwise, all layers are connected through \emph{rectified linear units}
(ReLu)~\cite{Krizhevsky:2012wl} as activation functions.
The linear layers of the whole model are all built out of 64 neurons.
For more information about (recurrent) neural networks see e.g.,~\cite{goodfellow2016deep}.

\subsubsection*{The Tactile Network}

\begin{figure}[!h]
    \centering{
    \includegraphics[width=0.75\linewidth]{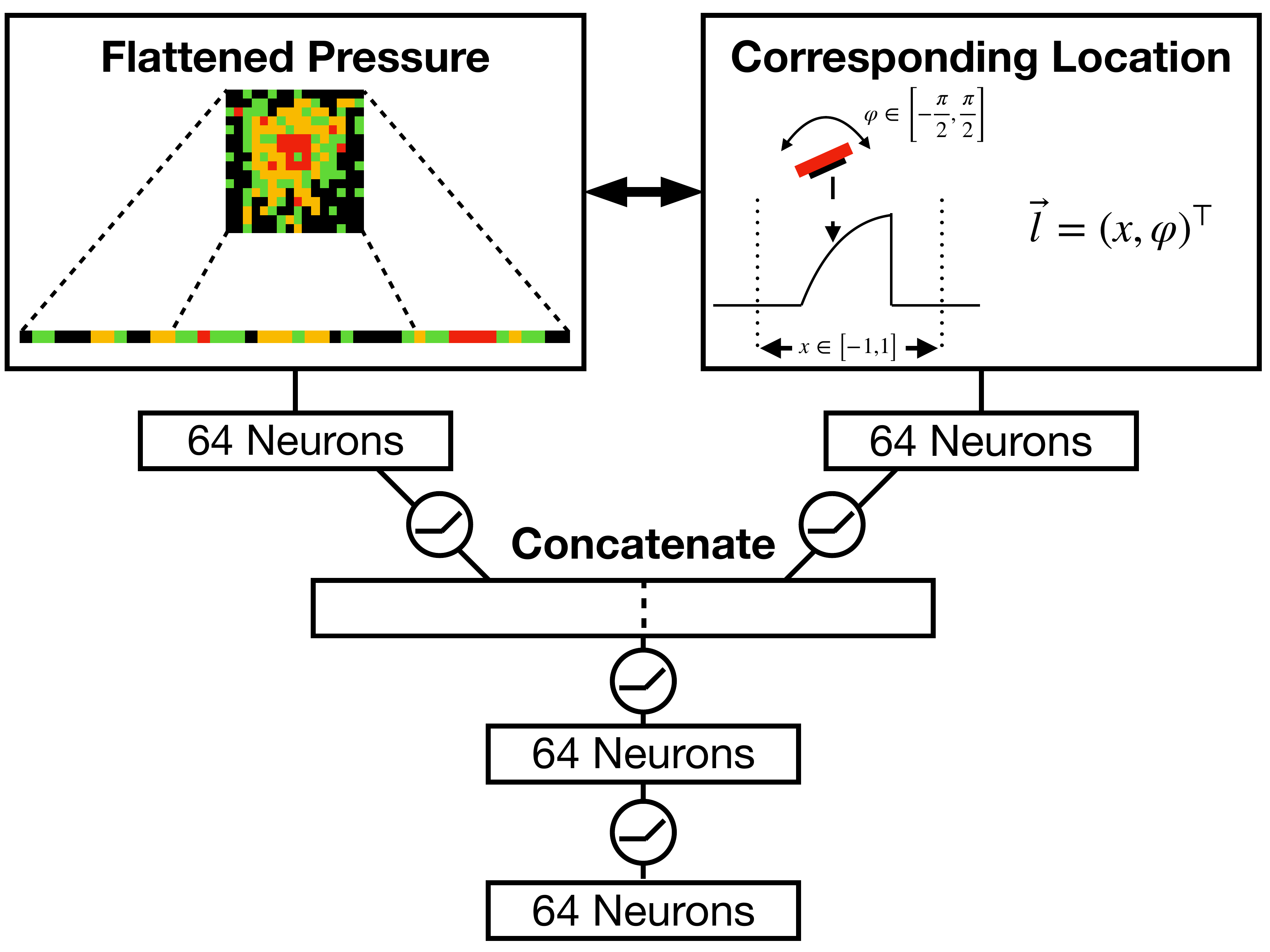}
}
\caption{{\bf Illustration of the Tactile Network}
The tactile network combines the normalized pressure $\vec{p}$ with the corresponding location $x$ and orientation
$\varphi$. Thus, the input vector for the pressure has the length $\mathrm{dim}(\vec{p})=256$ and the input vector for
the location-orientation pair $\mathrm{dim}(\vec{l})=2$ respectively. The small circles in-between the connections
indicate that the \emph{ReLu unit} is used as the activation function.} \label{fig:haptic_net}
\end{figure}

The \emph{tactile network} is displayed in detail in
Fig~\ref{fig:haptic_net}. It combines the
tactile response of the sensor $\vec{p}$ with the
corresponding location $x$ and angle $\varphi$. An important choice
is the approach used to combine \emph{what} (i.e., the pressure
$p$) with \emph{where} (i.e., position $x$ and orientation
$\varphi$). While \cite{Mnih:2014ti} use an element-wise
addition of the two features, \cite{Ba:2014ws,Larochelle:2010vo} suggest using
element-wise multiplication. In this work, based on
the performed tests, we concatenate the two resulting types of features
followed by two additional linear layers. In this way, we do not
impose  a specific inner structure on the combination process, but let
the network resolve this issue on its own.

\subsubsection*{The Location Network}
\begin{figure}[!h]
    \centering{
        \includegraphics[width=0.75\linewidth]{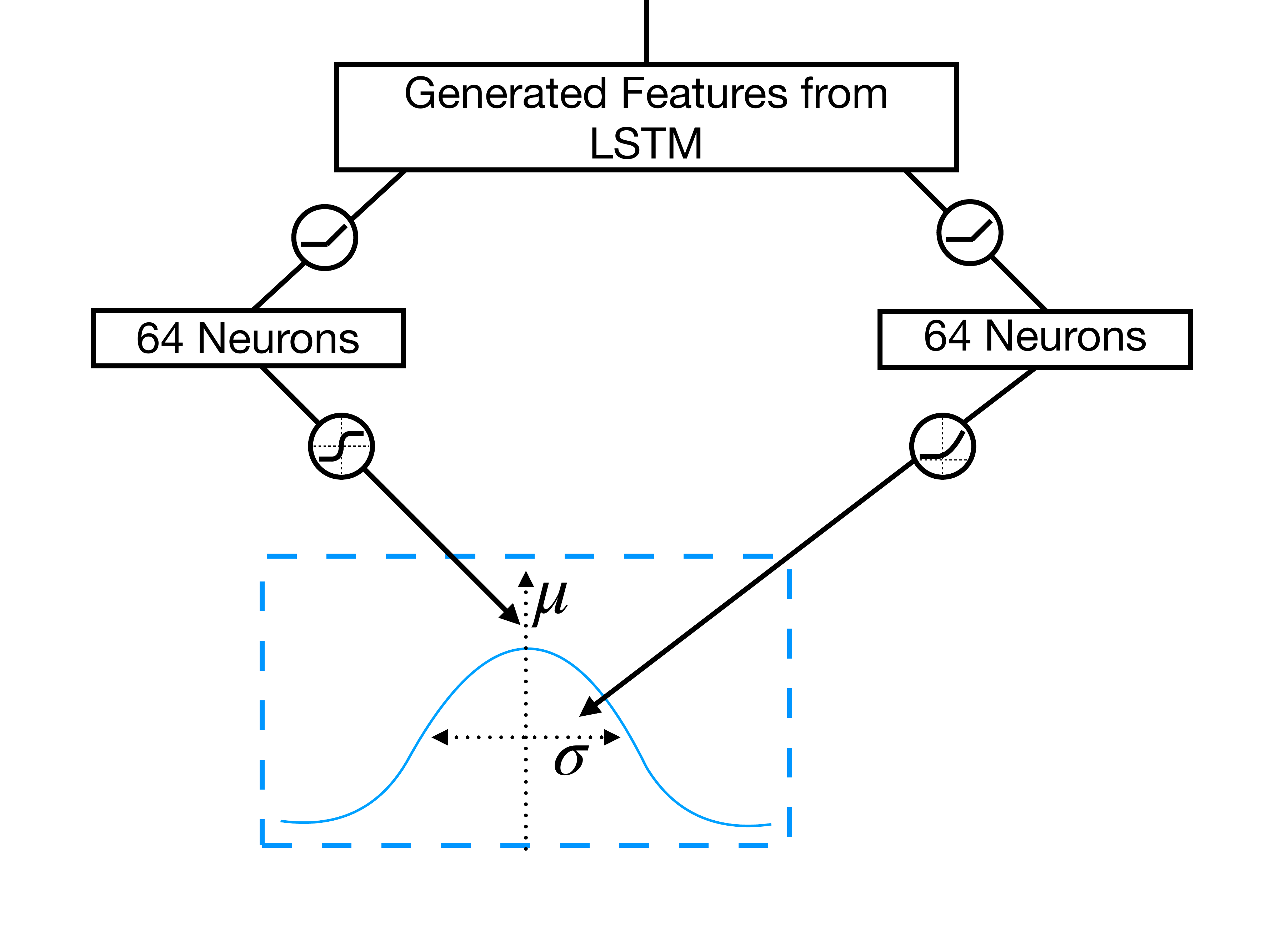}
}
\caption{{\bf Illustration of the Location Network}
The location network uses the generated features of the LSTM to determine a new location and
orientation (mean $\mu$, left branch) with the corresponding standard deviations ($\sigma$, right branch) for the tactile sensor using a stochastic policy.}
\label{fig:location_net}
\end{figure}

The \emph{location network} is designed to output the pose of the next
haptic glance. The feature vector that is used as the input to this module is the output that is generated by the LSTM
unit. It thus implicitly integrates shape information yielded by the previously performed glances. A stochastic location policy is modeled using two
Gaussian distributions for position and orientation, respectively %
with variable mean $\mu$ and standard deviation $\sigma$ as shown in Fig~\ref{fig:location_net}.

The features of the LSTM are propagated through a linear layer that
outputs the mean $\mu(\theta) \in \left[-1, 1 \right]$ and the
standard deviation $\sigma(\theta)$ of the Gaussian\footnote{
$\theta$ is referring to the corresponding weights of the model that are necessary to
generate the desired output, which is in this case $\mu$ or $\sigma$}.
The extent of exploration of the location policy is given by the size of the Gaussian's standard deviation $\sigma$.
While for large $\sigma$, the raw location of the glance, given by $\mu$, is imprecise, the location has more precision
for smaller $\sigma$.

The two above-mentioned pipelines are used for computing a distinct $\mu$ and $\sigma$ for the position and for orientation.
The used activation function for the output layers are chosen to limit the resulting values to a reasonable range. While
the \emph{tanh} is used as the activation function to generate the mean within the desired range,
the \emph{softplus function}~\cite{Dugas2001} is implemented as the activation function for the standard deviation. The output values
$\mu(\theta)$ and $\sigma(\theta)$ are then used to compute
the new location and orientation by sampling from 
the respective 1-dimensional Gaussians for each of the desired variables.

To ensure that the
location and position of the sensor remain within the predefined space
around the to-be-classified object and also that the orientation remains within
its boundaries, the sampled values of the Gaussians $\mathcal{N}(q;\mu,\sigma)$ are again
restricted to the range $q \in \left[-1, 1 \right]$. Thus, if $q$ is
sampled outside this range, it is resampled. The new pose vector is
then given as
\[
    \vec{l} = \left(q_x,q_\varphi \cdot \frac{\pi}{2} \right) = \left(x,\varphi \right).
\]

\subsubsection*{The Classification Network}

In order to classify a given object, the generated feature
vector of the LSTM is not only transferred to the \emph{location
  network}, but also propagated through a different linear layer that
  is then used for classification. To achieve this, the \emph{softmax-function} is
utilized  to encode the predicted class-affiliation of the
current object $o$ in a probability density $\pi(o|\vec{\tau}_{1:s};
\theta_t)$, representing the  current policy of the
reinforcement learning agent. Here, $\vec{\tau}_{1:S}(\theta_t)$ encodes
the accumulated LSTM feature vector after $S$ glances, using the
current set of weights $\theta_t$ at training step $t$. For classification, the class $o$
with the highest probability
\begin{equation}
    o = \argmax_{o'} \pi(o'|\vec{\tau}_{1:S};\theta_t)
    \label{eq:classify}
\end{equation}
is taken as the prediction.

\subsection*{Training}

For each training step, a
new batch of size $64$ is generated, where the to-be-classified
objects $o$ are uniformly chosen from the set of all four available objects.
The target loss function $\mathcal{L}$, used for training, is composed
of two different components: classification and location.  The update
rule for both parts is derived from the REINFORCE algorithm.
For the classification component of the loss,
we see the designed model as a reinforcement learner which has to
choose the right action in order to classify the given object. For
classifying the object correctly it receives a reward $r=1$, and $r=0$
otherwise. The predicted probability of correctly identifying the target
object $o$ after $S$ glances is then given as $\pi(o|\vec{\tau}_{1:S};\theta)$.
To this end, the \emph{categorical cross-entropy} can be used to compute the loss.

For learning the means $\mu_x$ and $\mu_\varphi$ of the location
component of the policy, the characteristic eligibility as outlined in
Eq~\eqref{eq:reinforce_mean} is used. $\sigma_x$ and $\sigma_\varphi$
are learned by applying Eq~\eqref{eq:reinforce_std}.
The hybrid update rule is then given by
\begin{equation}
  \notag \Delta_\Theta =- \alpha \cdot \left[\beta \cdot (r_t - b_t) \cdot \left( \zeta_{\mu_x} + \zeta_{\mu_\varphi} +
  \zeta_{\sigma_x} + \zeta_{\sigma_\varphi}\right) + \sum \limits_{o=0}^O \log(\pi(o)) \cdot y_o \right].
    \label{eq:update_weights}
\end{equation}
The function $\pi(o)$ gives the computed classification probability that the to-be-classified object is object $o$,
while $y_o$ is 1 if $o$ corresponds to the correct object and 0 otherwise.

The parameter $\beta$ controls the contribution of the different parts
of the update. While for $\beta=1$ both parts of the update contribute
equally to the weight update, a smaller factor of $\beta < 1$ assigns
more resources to the classification part. For $\beta=0$, the part of the update that
involves the location network is completely omitted~\cite{Larochelle:2010vo}.

The baseline layer is updated separately, using the mean-squared error.
Instead of training the baseline only on the accumulated tactile information of the last
glance $\vec{\tau}_{1:S}$, the training can be improved by also using all included 
sub-sequences $\vec{\tau}_{1:s}$ with $s \leq S$~\cite{Larochelle:2010vo}.
This leads to the loss function
\begin{eqnarray}
    \mathcal{L}_b = \sum \limits_{s=1}^{S} \left[ R_s - b(\vec{\tau}_{1:s}; \theta_b)\right]^2.
    \label{eq:baseline}
\end{eqnarray}

The overall network model is trained using stochastic gradient descent with
Nesterov momentum~\cite{Nesterov:1983wo,Nesterov:2013wi}. The chosen
learning rate of $\alpha_0$ decays towards $\alpha_\mathrm{min}$ every
training-step $t$ with a decay factor of $\delta_\alpha$ and a
step-size of $T$ according to
\begin{equation*}
\alpha_t = \max \left(\alpha_\mathrm{min}, \alpha_0 \cdot \delta_\alpha^{\frac{t}{T}} \right).
\end{equation*}

Due to the design of the network that generates a location for the
next haptic glance, no fixed training set can be used to train the
classifier. The current batch specifies only the to-be-classified
objects, while the first pressure-location pair is chosen by the first
random glance for each object. The location for any further glance is
chosen by the current state of the  location policy of the network.

\section*{Experiments}

To perform an empirical examination of the validity of the network
architecture, we perform a series of evaluations with a focus on each
one of the three modules: the LSTM, the location network, and the
tactile network. The core of the evaluation approach focuses on
the recurrent LSTM unit that plays a central role in feature
extraction and integration.  Our hypothesis is that by employing LSTM
we increase both the classification accuracy and the efficiency of the
pose control.  To test the efficiency of the LSTM on both tasks, the
classification accuracy is computed while training the network on a
varying number of glances.  In addition to the final classification
accuracy, the individual classification accuracies after each glance
are evaluated. To demonstrate the efficiency of using a recurrent unit
instead of a simple linear hidden layer, the experiment is repeated
with the LSTM replaced by a linear layer of the same size (i.e., 256 neurons).

The second part of the evaluation is dedicated to the pose control by
the location network.  We evaluate it during the learning
process, and compare the results against a model with a random
location choice. To this end, we omit the location network and provide
the model with new locations $x \in \left[-1,1\right]$ and
orientations $\varphi \in \left[-\pi/2, \pi/2 \right]$ that are
sampled from a uniform distribution. For training, only the
classification part of Eq~\eqref{eq:update_weights} is used to create
the weight update, while $\beta$ is set to $0$.

In the third part of the evaluation, the different approaches for
combining the tactile information with its corresponding location
(What \& Where) are compared.

All models are trained for $50 \cdot 10^3$ steps.
In order to measure the performance after a certain number of training
steps, the training is stopped. This is followed by estimation of the mean
classification accuracy of $100$ newly generated batches, using the
currently available policy.
In our experiments, the ``classification accuracy'' or ``classification
performance'' is defined as the
probability of the model to correctly classify the current object.
To obtain a statistically correct
measure of the accuracy, each experiment is repeated 10 times.  For
the final evaluation, the mean accuracy of these experiments is
computed with the standard deviation of the mean as the accuracy measure error.

\subsection*{Hyperparameters}
Table~\ref{tab:hyperparams} lists the hyperparameters that are used
for all experiments.  The parameters are chosen according to
\emph{random search}~\cite{Bergstra:2012ux} with a fixed number of 3
glances, followed by additional manual tuning. The weights of all
layers are initialized using \emph{He normal initialization}~\cite{He:2015vx}
with a bias of $0$.

\begin{table}[!ht]
\centering
\caption{
{\bf Hyperparameters employed for network training}}
\begin{tabular}{|l|l|}
\hline
\multicolumn{1}{|l|}{\bf Parameter} & \multicolumn{1}{|l|}{\bf Value}\\ \thickhline
Batch Size & $64$ \\ \hline
Initial learning rate $\alpha$ & $8 \cdot 10^{-4}$ \\ \hline
Learning rate decay factor $\delta_\alpha$ & $0.97$ \\ \hline
Learning rate update step-size $T$ & $800$ \\ \hline
Minimal learning rate $\alpha_0$ & $10^{-6}$ \\ \hline
Used optimizer & SGD with Nesterov momentum \\ \hline
Momentum & $0.9$ \\ \hline
Location weight $\beta$ & $0.4$ \\ \hline
Hidden state size - LSTM & $256$ neurons \\ \hline
Size of linear layers& $64$ neurons \\ \hline
\end{tabular}\label{tab:hyperparams}
\end{table}

\subsection*{Creation of the dataset}

In order to perform quick optimization and testing, we conducted
multiple experiments on a pre-recorded dataset $\mathcal{D}_o$ (see
\nameref{S1_Dataset})
generated in Gazebo, previous to the experimental runs, for each object $o$.
The dataset contains tuples $d_o = (x, \varphi, \vec{p})$. Here
$\vec{p}$ is the normalized pressure-vector $\vec{p}$,
$x \in \left[-1,1 \right]$ the respective position of the sensor
within the location space and $\varphi \in \left[-\pi/2,\pi/2 \right]$
 the angle.  For each object the recording of the tuples $d_o$ starts
with the position $x=-1$ and the orientation $\varphi=-\pi/2$. These
two parameters are then both incremented with a step size of $\Delta_x
= 0.01$ and $\Delta_\varphi = \pi \cdot 0.01$, leading to $201 \times
201$ prerecorded data-points $d_o$ for each object. The complete
dataset has then a size of roughly $161 \cdot 10^3$ data-points that
can be picked to approximate the sensor pose generated by the location
network. For a new pair $(x, \varphi)$ generated by the network, the
closest data-point $d_o$ is selected from the pre-recorded data set.

\section*{Results}

The main results of the conducted experiments are summarized in Table~\ref{tab:results}. It
displays the classification accuracies for all three variants of the
architecture as described above and shows the corresponding
results for an increasing number of glances.
The full meta-controller model $\pi_{\mathcal{M}}$ contains all trained components including the LSTM module and the
location network. The random location policy approach $\pi_\mathrm{rloc}$ substitutes the location network with a
random location generator. $\pi_\mathrm{MLP}$ substitutes the LSTM unit with a linear
layer of the same size with a ReLu as its activation function. As the neural network is now built out of linear layers only, it can be seen as a \emph{multi-layer perceptron} (MLP). In the
last column, labeled $\left< \pi_\mathrm{MLP} \right>$, the classification performance of $\pi_\mathrm{MLP}$ is evaluated by averaging over all conducted glances.

The ``full
model'' $\pi_{\mathcal{M}}$ (see column 1) reaches a classification accuracy of about $99.4\%$ on the
pre-recorded dataset. While the accuracy using one random glance is
only $\approx 55\%$, it continuously improves when more glances can be
executed. Granting the model just one more glance leads to an accuracy
of about $83\%$. Overall, accuracy improvement for the full model is
faster than for the other two tested architectures, up to its
convergence after about 6 glances are performed.


\begin{table}[!ht]
\centering
\caption{
{\bf Best Classification Performance for the Different Number of Glances}}
\begin{tabular}{|l|l|l|l|l|}
\hline
\multicolumn{1}{|l|}{\bf \# Glances} & \multicolumn{1}{|l|}{\bf 1.  $\pi_{\mathcal{M}}$} & \multicolumn{1}{|l|}{\bf 2. $\pi_\mathrm{rloc}$} & \multicolumn{1}{|l|}{\bf 3. $\pi_\mathrm{MLP}$} & \multicolumn{1}{|l|}{\bf 4. $\left< \pi_\mathrm{MLP} \right>$}\\ \thickhline
1 & $0.547 \pm 0.002$ & $0.547 \pm 0.002$ & $0.570 \pm 0.003$ & $0.570 \pm 0.003$ \\ \hline
2 & $\mathbf{0.831 \pm 0.002}$ & $0.753 \pm 0.002$ & $0.668 \pm 0.003$ & $0.645 \pm 0.003$ \\ \hline
3 & $\mathbf{0.910 \pm 0.001}$ & $0.858 \pm 0.001$ & $0.662 \pm 0.003$ & $0.829 \pm 0.002$ \\ \hline
4 & $\mathbf{0.942 \pm 0.001}$ & $0.917 \pm 0.001$ & $0.662 \pm 0.003$ & $0.912 \pm 0.002$ \\ \hline
6 & $\mathbf{0.977 \pm 0.001}$ & $0.971 \pm 0.001$ & $0.670 \pm 0.004$ & $0.976 \pm 0.001$ \\ \hline
8 & $0.988 \pm 0.001$ & $0.990 \pm 0.000$ & $0.668 \pm 0.004$ & $0.993 \pm 0.000$ \\ \hline
10 & $0.994 \pm 0.001$ & $0.995 \pm 0.000$ & $0.668 \pm 0.003$ & $0.997 \pm 0.000$ \\ \hline
\end{tabular}
\begin{flushleft} The table lists the best measured classification performance after $50 \cdot 10^{3}$ training steps for the different tested model variants.
\end{flushleft}
\label{tab:results}
\end{table}

Column 2 presents the results of the
random location policy. It starts from the same performance as the full model
(since the first glance is random in both policies) and from there approaches
its asymptotic performance more slowly, making its performance inferior when
only 2 to 6 glances can be invested.
Thus, our model is able to learn to efficiently extract
important information when the number of possible interactions with
the given object are limited.

If the recurrent LSTM unit is replaced with a linear layer of the same
size (column 3), the classification accuracy does not rise beyond
$67\%$, constituting the worst result. Due to missing recurrent
connection, and the fact that the accuracy is evaluated only after the
last glance, the MLP-based architecture $\pi_\mathrm{MLP}$ is optimized based only on the
last glance, and therefore does not improve after two glances.

However, by averaging its output according to
\[
    o = \argmax_{o'} \frac{1}{S} \sum \limits_{s = 1}^S \pi(o'|\vec{\tau}_{s:S};\theta_t)
\]
the performance of this averaged MLP model becomes very similar to the random model (column 2). Asymptotically (here: ten or more glances), all except the MLP model reach practically perfect classification.

\begin{figure}[!h]
    \centering{
        \includegraphics[width=0.75\linewidth]{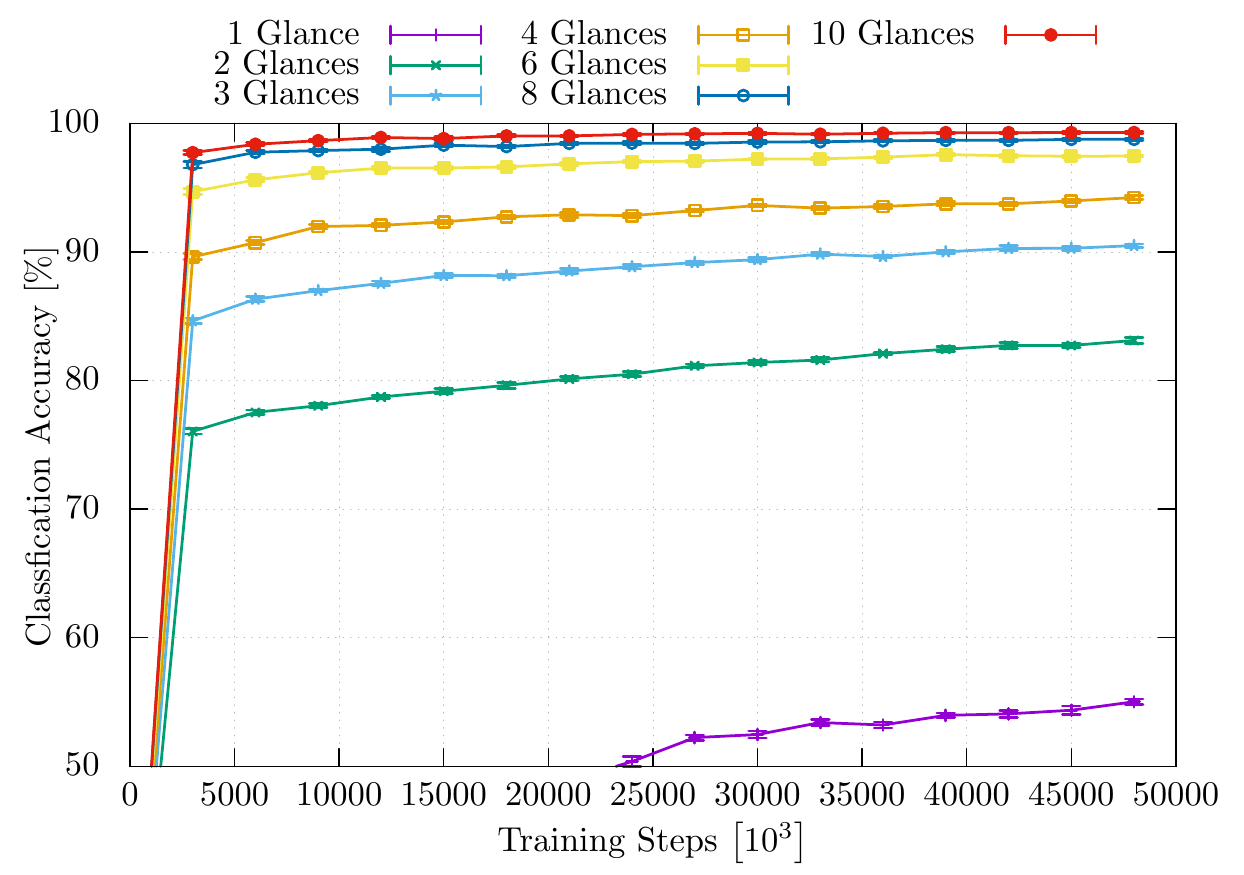}
    }
\caption{{\bf Classification Accuracy} The time course of the classification accuracy during the training is visualized for the model while it is trained to classify using a different number of glances.}
\label{fig:classification_performance}
\end{figure}

\begin{figure}[!h]
    \centering{
        \includegraphics[width=0.75\linewidth]{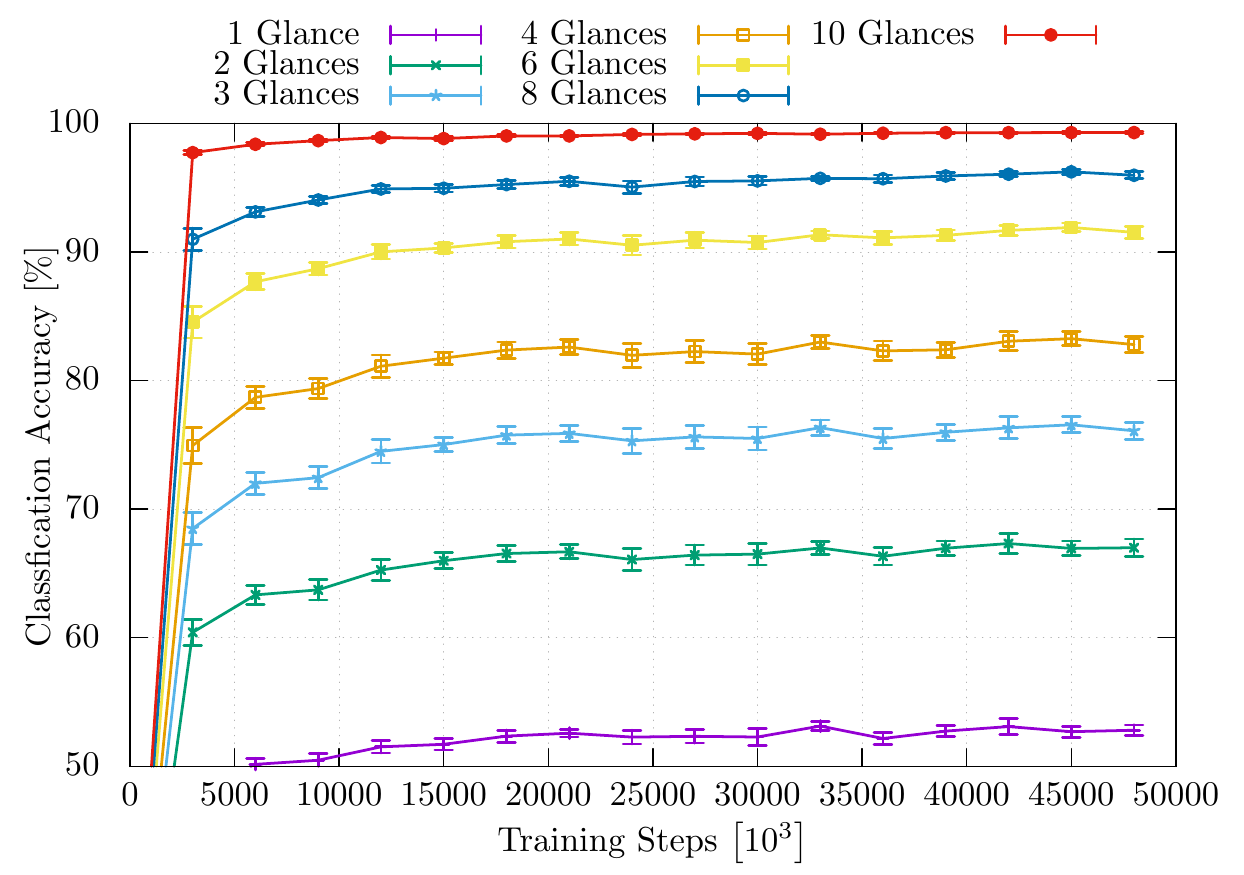}
}
    \caption{{\bf Classification Accuracy} Classification
    accuracy of the individual glances in one classification event using the LSTM
layer is displayed. The classification event uses 10 glances to classify each object.}
\label{fig:classification_performance_glances}
\end{figure}

Fig~\ref{fig:classification_performance} shows the time course of learning 
of the model for the different numbers of performed
glances.
Additionally, the individual classification accuracy for each glance
within one classification event that uses 10 glances is visualized in
Fig~\ref{fig:classification_performance_glances}.
The accuracy of the individual glances within a
classification event differs from the ones in
Fig~\ref{fig:classification_performance}.

\begin{figure}[!h]
    \centering{
        \includegraphics[width=0.75\linewidth]{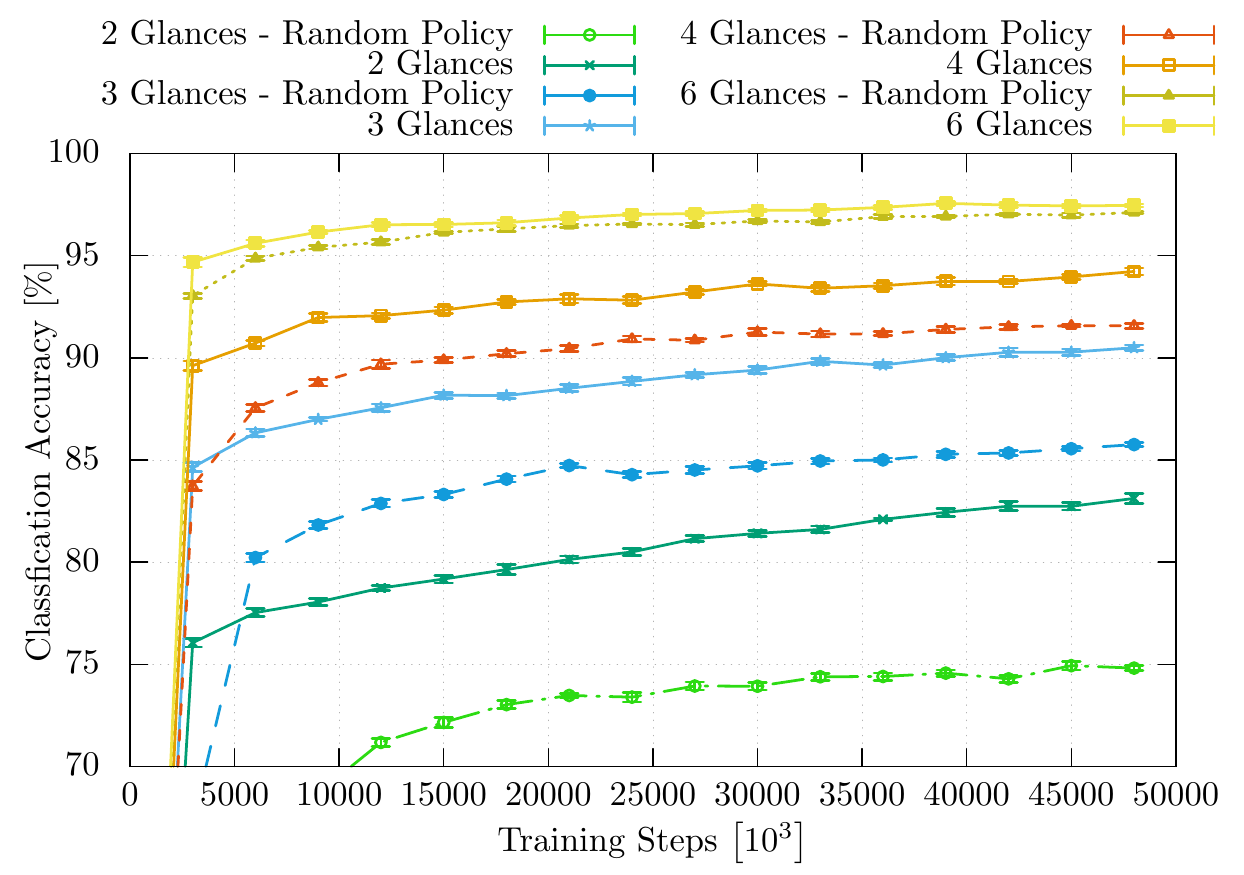}
}
\caption{{\bf Classification Accuracy} Comparison of classification accuracies between the model that relies on a learned
location policy $\pi_\textrm{loc}$ (solid lines) and the model that
uses a random location policy (dashed lines).}
\label{fig:LSTM_vs_RAND}
\end{figure}

Fig~\ref{fig:LSTM_vs_RAND} presents a detailed performance comparison
between the $\pi_\mathrm{rloc}$ and the full model $\pi_\mathcal{M}$.
Here, one can again see that a huge performance gap exists when the
model is able to execute only a small number of glances and that this
gap is progressively closed as the number of glances is increased.
Fig~\ref{fig:LSTM_vs_RAND} shows that the impact of the learned
location policy is more visible when the model is trained on a smaller
number of glances.  The model $\pi_\mathcal{M}$  learns to classify
objects based on limited information more efficiently. 

Table~\ref{tab:results_what_where} lists the best classification
accuracies of the model using 3 glances for the different ways of combining the
normalized pressure vector $\vec{p}$ with the corresponding location
$\vec{l}$.
The procedure to combine the two sets of features via
concatenation and then processing the result through two layers gives slightly better
results than the element-wise addition, but
clearly outperforms the approaches of element-wise multiplication and the concatenation
approach using one layer.

\begin{table}[!ht]
\centering
\caption{
{\bf Learning Performance: What \& Where}}
\begin{tabular}{|l|l|}
\hline
\multicolumn{1}{|l|}{\bf Combiner} & \multicolumn{1}{|l|}{\bf Best Performance}  \\ \thickhline
elem. multiplication & $0.873 \pm 0.002$  \\ \hline
concat. followed by 1 layer & $0.899	\pm 0.002$ \\ \hline
elem. addition & $0.902 \pm 0.001$ \\ \hline
\textbf{concat. followed by 2 layers} & $\mathbf{0.905 \pm 0.001}$ \\ \hline
\end{tabular}
\begin{flushleft} The table lists the best measured classification performance within the $50 \cdot 10^{3}$ training steps for the different tested approaches of combining the normalized pressure $\vec{p}$ with the location-orientation pair $(x,\varphi)$.
\end{flushleft}
\label{tab:results_what_where}
\end{table}


\section*{Discussion}

The performed evaluations demonstrated that the proposed model is able
to classify the objects with an accuracy of nearly $100\%$ by actively
acquiring an optimized sequence of tactile sensor measurements. In
this approach the data is generated on-the-fly by haptic interaction
with the environment, performed by means of haptic glances and
directed by the history of previous tactile events.  The results of
the conducted experiments show that the full network architecture
$\pi_\mathcal{M}$, including the recurrent LSTM network and the
location network, is capable of controlling the execution of haptic
glances in the most efficient way. The architecture performs better
with a trained location network than with a random location policy
$\pi_\mathrm{rloc}$.  Employing the LSTM to represent the sequence
history yields better performance in comparison to the memory-less
architecture $\pi_\mathrm{MLP}$. Here, the location network only
slightly improves the location w.r.t. the task-relevant information
with the increasing number of glances. In comparison with
$\pi_\mathcal{M}$, a  good but less efficient performance of the
$\left< \pi_\mathrm{MLP} \right>$ that accumulates individual
classification decisions may be due to the averaging out of noise with
the increasing number of glances.  Therefore, both recurrence and an
optimized location control are likely to be necessary ingredients of
an efficient haptic exploration model in our scenario.
These results may be constrained by the simplicity of the 3D shapes
considered in the experiment.
For an extensive evaluation of the proposed approach, the creation of data sets
with a greater number of different objects would be necessary,
including stimuli that are more challenging to differentiate without
an optimized control strategy.
For the described case, we expect that the efficiency and
accuracy trends would become more evident.

The network architecture $\pi_\mathcal{M}$ merely fuses and
accumulates the data, whose representation is optimized with the goal
to achieve the most accurate and efficient execution for a given
task. Therefore, it provides a general interface, which has a capacity
to accommodate for different types of haptic glance parameterizations.
However, our approach to parameterization and control was deliberately
very rudimentary in this work.  The currently employed
minimalistic haptic glance is controlled by a one-dimensional
translation and rotation, characterized by a uni-variate pressure
output.  This simplification was coupled to the experimental design
targeting exploration of one-dimensional curvature features.  Other
types of haptic glance controller parameterizations are desirable, in
case other features than the curvature need to be explored. On the one
hand, both the number, type of the control parameters and the outputs
are very likely to be determined bottom-up by the features of the 3D
environment in which haptic interaction is performed. On the other
hand, they are determined in a top-down fashion by the task of the interaction.  It
remains an open question how to automatically derive the minimal
parameterization and the output of the haptic glance controller
depending on the features of the environment, the task, the available
degrees of freedom of the employed device and its tactile
capabilities.

The modularity of our model should, however, provide the functionality
to adapt to more complex sensor devices as different modules of the
HAM just have to be extended to cope with the increasing number of
control dimensions.  In the current configuration, our model has a
total number of 741248 trainable weights.  Within the simplest case,
for each additional trainable parameter that is provided to the
low-level haptic glance controller an additional output stream with at
least one additional linear layer (with e.g. 64 neurons) has to be
added to the location network. While this procedure does not
necessarily  lead to a significant increase in the number of
trainable weights, too many additional control parameters might exceed
the memory and processing capacity of the  LSTM network.
The LSTM network contains --- in contrast to the linear layers --- a large
amount of the trainable weights.  A necessary amplification of its
size or the addition of a second LSTM network in order to increase 
performance might  have a higher impact on the model's size and
 its training time.  While the designed hybrid loss might be a
reasonable approach when only two control parameters have to be
adapted, a higher number could slow down the convergence process of
the model. One possible way out of this dilemma might be to separately
train the classification and control part of the HAM with different
loss functions that share the achieved reward.

\section*{Conclusion and Future Work}

In this work we have proposed the first implementation of a
controller, inspired by the concept of \textit{haptic glances}.
Provided a pose parameter as an input, a floating tactile sensor array
touches the surface at the specified location and yields the resulting
pressure vector. We have trained a meta-controller network
architecture to perform an efficient haptic exploration of 3D shapes
by optimally parametrizing the haptic glance controller to perform a
sequence of glances and identify 3D objects. Tests of the architecture
have been successfully performed in a physics-driven simulation
environment.

The structure of the meta-controller includes a mechanism that
accumulates the data acquired during execution of the task and
parameterizes the future haptic glances based on the optimized
representation of this data. However, the current mechanism
performing this temporal integration --- based on an LSTM and inspired by
the functionality of the working memory --- may not be sufficient for an
execution of a more complex task consisting of multiple task stages,
such as e.g. haptic search, or a contact-rich object manipulation. In
such tasks, it may be necessary to save the representation of data
existing in the working memory to a long-term memory, from which this
information could be retrieved at a later stage in the task execution.
To this end, the meta-controller needs to communicate with an extra
structure, based on e.g. hashing, such as Neural Turing
Machine~\cite{Graves2014NTM} to access features acquired at multiple
previous time slots during interaction with the target topology.

To support our claim that the resulting policy can enable a robot
equipped with a tactile sensor to perform efficient object
identification by touch, we see performing tests with a (simulated) robot
platform, equipped with a Myrmex tactile sensor array as our next
task. Furthermore, we will extend our experimental design
with the second curvature dimension and, corresponding to this, an
extra degree of freedom in our haptic glance controller.

Due to the fact that the pose is
sampled from a Gaussian distribution, it is highly unlikely that the
same position or orientation is sustained during the
exploration. Therefore, the current approach results in a jumpy
energy-inefficient exploration trajectory which makes a more
energy-efficient policy desirable.  Consequently, the meta-controller
optimization should be extended to enable a smoother trajectory
generation. This may be possible by a careful shaping of the reward
function or a further refinement of the location network.

Beyond performing haptic object identification, we believe that the
developed procedure may be applied to enable a robot to perform
complex manipulation tasks that heavily rely on haptics.  Execution of
a more complex tasks such as above-mentioned haptic search commonly
involve multiple types of strategies, targeting exploration of
different types of haptic features, e.g. movability or rigidity.  This
may be possible by implementing a set of low-level haptic glance
controllers characterized by different parameterizations and outputs
accompanied by a gating mechanism that enables the overall model to
switch between them.

\section*{Supporting Information}

\paragraph*{S1 Code.}
\label{S1_Link_Gazebo}
{\bf Gazebo} The simulation software is available under the following link: \\ \url{http://gazebosim.org/}

\paragraph*{S2 Code.}
\label{S2_Link_Myrmex}
{\bf Myrmex Simulation} Code of the tactile simulation is available under the
following link: \\
  \url{https://github.com/ubi-agni/gazebo_tactile_plugins}

\paragraph*{S3 Code.}
\label{S3_Link_Ros}
{\bf ROS} The software is available under the following link: \\ \url{http://www.ros.org/}

\paragraph*{S4 Code.}
\label{S4_Link_HoG}
{\bf The ``Hand of God`` Plugin} The plugin is available under the following link: \\
\url{https://github.com/ros-simulation/gazebo_ros_pkgs/blob/kinetic-devel/gazebo_plugins/src/gazebo_ros_hand_of_god.cpp}

\paragraph*{S5 Code.}
\label{S5_HAM}
{\bf The Haptic Attention Model} 
A link to the source code can be found at: \\ \url{http://doi.org/10.4119/unibi/2936475}

\paragraph*{S1 Video.}
\label{S1_Video}
      {\bf Contact Information in Gazebo  } A visualization of the contact information in Gazebo.

\paragraph*{S2 Video.}
\label{S2_Video}
{\bf Modular Haptic Stimulus Board (MHSB)} The video presents design and applications with MHSB \url{https://www.youtube.com/watch?v=CftpCCrIAuw}

\paragraph*{S3 Video.}
\label{S3_Video}
      {\bf Gazebo Simulation --- Haptic Glance Controller}
      The video describes the experimental setting and illustrates the functionality
      of the haptic glance controller during object exploration.

\paragraph*{S1 Project.}
\label{S1_Project}
{\bf Modular Haptic Stimulus Board  (MHSB)} Project web-site is available under the
following link: \url{https://ni.www.techfak.uni-bielefeld.de/node/3574}

\paragraph*{S1 Dataset.}
\label{S1_Dataset}
{\bf Recorded Dataset of Glance Locations and the Corresponding Pressure Data}
The recorded dataset is available at: \\
\url{http://doi.org/10.4119/unibi/2936475}

\begin{figure}[!ht]
    \centering{
        \includegraphics[width=0.85\linewidth]{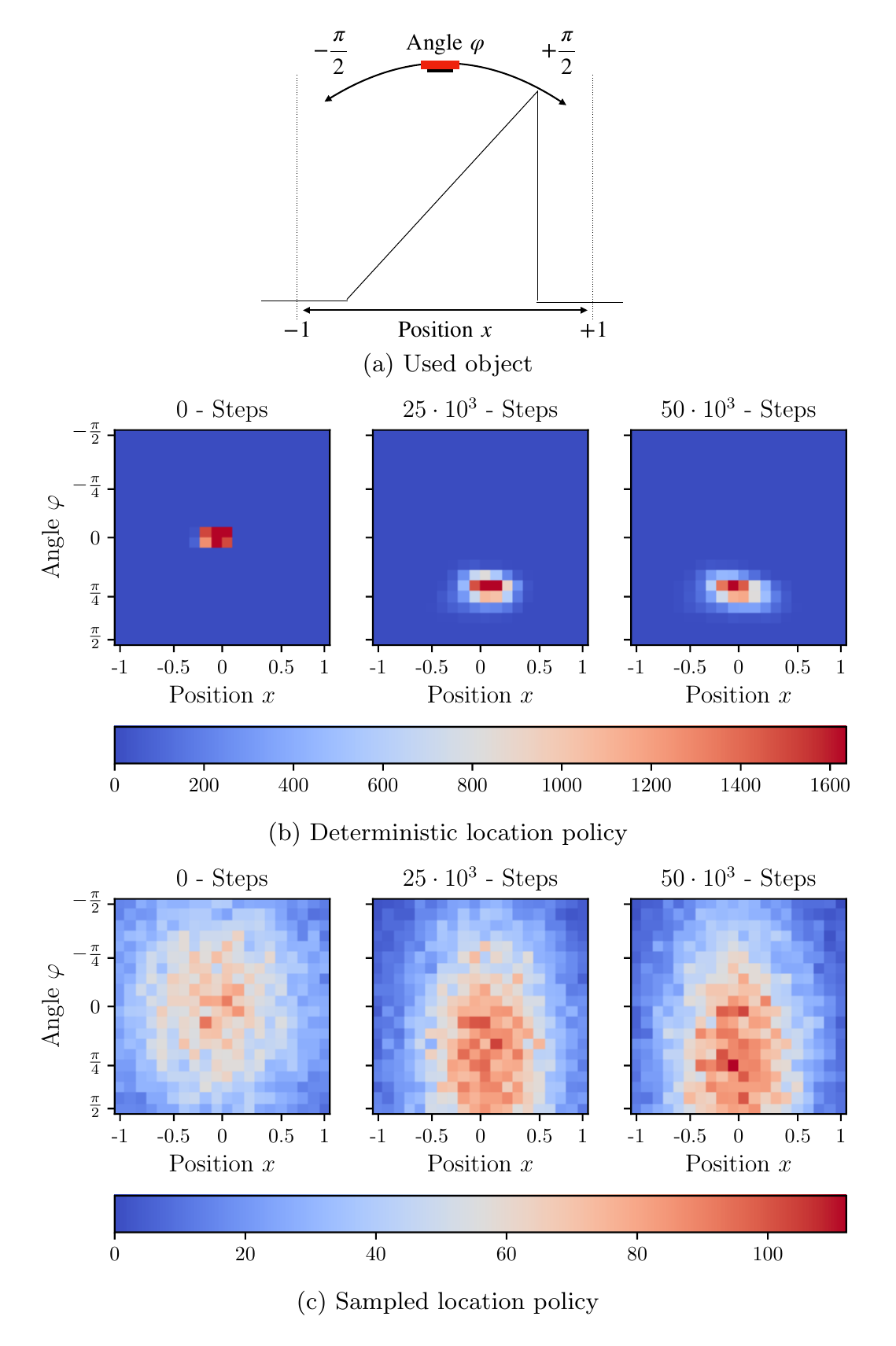}
    }
\caption{{\bf Learning Where to Touch} Illustration of the shaping process of the location policy for the sketched object illustrated in (a). Figure (b)-(d) are visualizing the learned $\mu_x$ and $\mu_\varphi$. Figure (e)-(g) are showing the corresponding sampled positions for the last glances at the given training-step.}
\label{fig:heatmap_absolute_2}
\end{figure}

\begin{figure}[!ht]
    \centering{
        \includegraphics[width=0.85\linewidth]{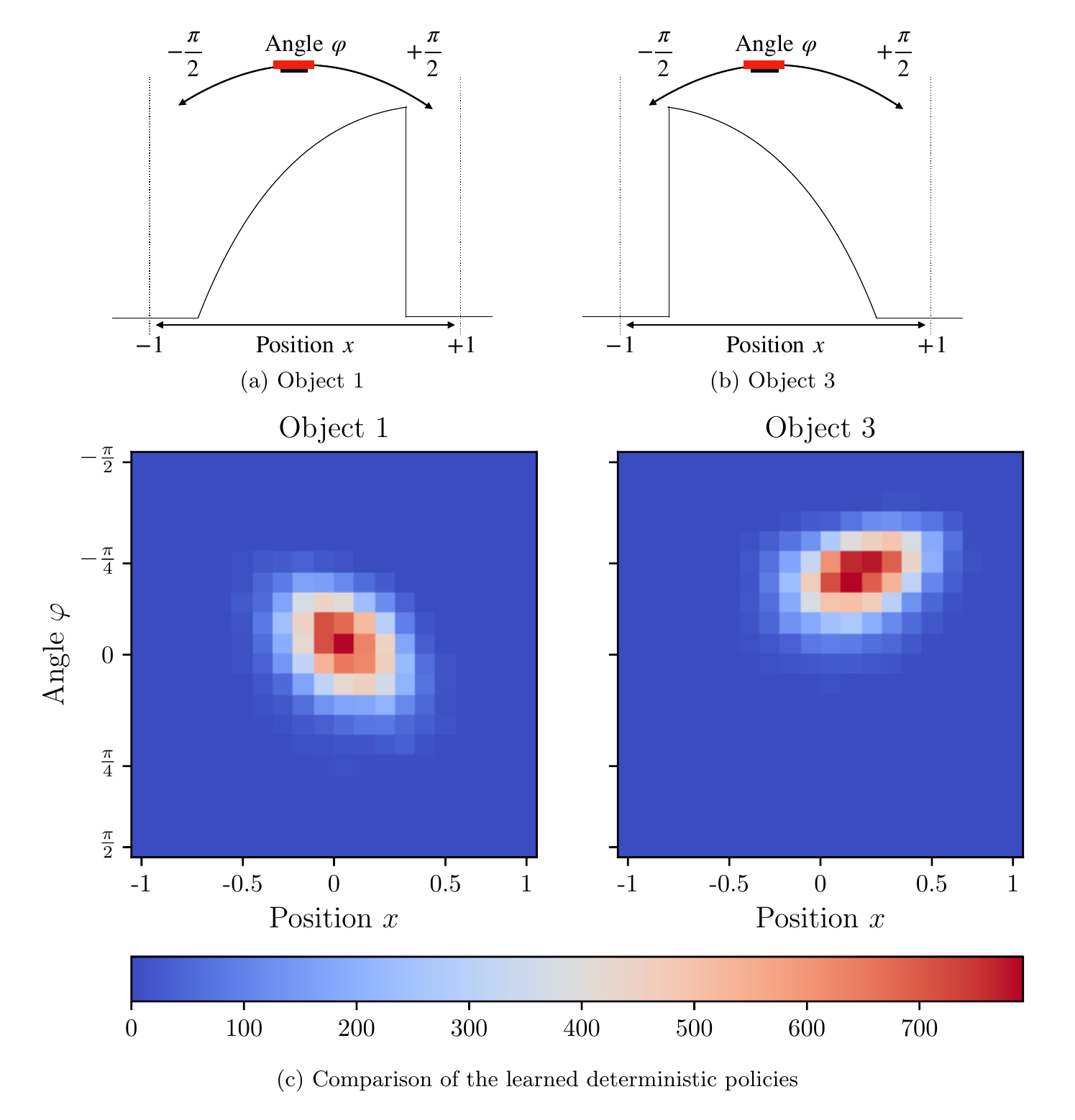}
    }
\caption{{\bf Learning Where to Touch} Illustration of the learned deterministic policy
for the sketched objects illustrated in (a) and (b) after $50 \cdot 10^3$
training steps. Figure (c) and (d) are visualizing the learned policy of the means $\mu_x$ and $\mu_\varphi$ for the last glances.}
\label{fig:heatmap_absolute_1_3}
\end{figure}

\paragraph*{S1 Appendix.}
\label{S1_Appendix}
{\bf Heat-Map Visualization of Learning Good Locations}
To visualize the learning of ``good locations'' for tactile
classification of the objects, heat-maps are created during the learning of
model that is trained on 10 glances.
The heat-maps show how
often a specific location-orientation pair was visited during the
classification process of the performance runs. For this purpose, the
location-orientation space was discretized in $20 \times 20$
bins. To generate the location-orientation profiles for
the objects, $1000$ batches are evaluated in each performance run.
 During the performance run, the number of visits to the
different bins was counted for the last executed glance for each individual classification.
The results are illustrated in Fig~\ref{fig:heatmap_absolute_2} and Fig~\ref{fig:heatmap_absolute_1_3}.
To create the heat-maps, the bin with the most visits is identified. This number of visits is then taken as the maximum value to rank the $400$ bins according to the number of visits.
Fig~\ref{fig:heatmap_absolute_2} shows the location policy for the triangular-shaped object. Fig~\ref{fig:heatmap_absolute_2}~(e)-(g) illustrate the evolution of the learned means $\mu_x$ and $\mu_\varphi$ of the location policy. While the generated values are centered around $x=0$ and
$\varphi=0$ at the beginning of the training, the prioritized angle changes to $\varphi \approx \pi/4$ during the
learning process. Fig~\ref{fig:heatmap_absolute_2}~(b)-(d) illustrate the corresponding sampled policy.

The learned location-policy of the model differs between the different to-be-classified objects. While the best location policy for the triangular-shaped object (Fig~\ref{fig:heatmap_absolute_2}) seems to be a plateau at $\varphi=\pi/4$ around $x=0$, the location policies for the objects in Fig~\ref{fig:heatmap_absolute_1_3} tries to cover a broader range of different angles. It is also worth to mention that the symmetry of the two illustrated objects in Fig~\ref{fig:heatmap_absolute_1_3} is also reflected within the learned location policy.

\section*{Acknowledgments}
This
research/work was supported by the Cluster of Excellence Cognitive
Interaction Technology 'CITEC' (EXC 277) at Bielefeld University,
which is funded by the German Research Foundation (DFG).
We also wish to acknowledge the technical support provided by Guillaume Walck.

\clearpage

\bibliographystyle{plos2015}

\end{document}